\documentclass[10pt,twocolumn,letterpaper]{article}

\usepackage{cvpr}
\usepackage{times}
\usepackage{epsfig}
\usepackage{graphicx}
\usepackage{amsmath}
\usepackage{amssymb}

\usepackage[pagebackref=true,breaklinks=true,letterpaper=true,colorlinks,bookmarks=false]{hyperref}

\cvprfinalcopy 

\def\cvprPaperID{306} 

\makeatletter
\newcommand\footnoteref[1]{\protected@xdef\@thefnmark{\ref{#1}}\@footnotemark}
\makeatother
\ifcvprfinal\fi
\begin{document}
	
	\title{Large Scale Incremental Learning}
	
	\author{Yue Wu$^1$\quad Yinpeng Chen$^{2}$\quad Lijuan  Wang$^{2}$\quad Yuancheng  Ye$^{3}$\\ Zicheng  Liu$^{2}$\quad Yandong  Guo$^{2}$ \quad Yun Fu$^{1}$\\
		$^1$Northeastern University\quad $^2$Microsoft Research\quad  $^3$City University of New York\\
		{\tt\small \{yuewu,yunfu\}@ece.neu.edu, yye@gradcenter.cuny.edu}\\ {\tt\small\{yiche,lijuanw,zliu\}@microsoft.com, yandong.guo@live.com} 
	}
%
	
	\maketitle
	
	\begin{abstract}
		Modern machine learning suffers from \textit{catastrophic forgetting} when learning new classes incrementally. The performance dramatically degrades due to the missing data of old classes. Incremental learning methods have been proposed to retain the knowledge acquired from the old classes, by using knowledge distilling and keeping a few exemplars from the old classes. However, these methods struggle to \textbf{scale up to a large number of classes}. We believe this is because of the combination of two factors:  (a) the data imbalance between the old and new classes, and (b) the increasing number of visually similar classes.  
		Distinguishing between an increasing number of visually similar classes is particularly challenging, when the training data is unbalanced.
		We propose a simple and effective method to address this data imbalance issue. We found that the last fully connected layer has a strong bias towards the new classes, and this bias can be corrected by    a linear model. With two bias parameters, our method performs remarkably well on two large datasets: ImageNet (1000 classes) and MS-Celeb-1M (10000 classes), outperforming the state-of-the-art algorithms by 11.1\% and 13.2\% respectively.
	\end{abstract}
	
	\section{Introduction}
	
	
	Natural learning systems are inherently incremental where new knowledge is continuously learned over time while existing knowledge is maintained ~\cite{rebuffi2016icarl,li2016learning}. Many computer vision applications in the real world require incremental learning capabilities. For example, a face recognition system should be able to add new persons without forgetting the faces already learned. However, most deep learning approaches suffer from \textit{catastrophic forgetting} ~\cite{McCloskey-Cohen-PLM-1989}  - a significant performance degradation, when the past data are not available. 
	
\begin{figure}[!t]
		\begin{center}
			\includegraphics[width=0.6\linewidth]{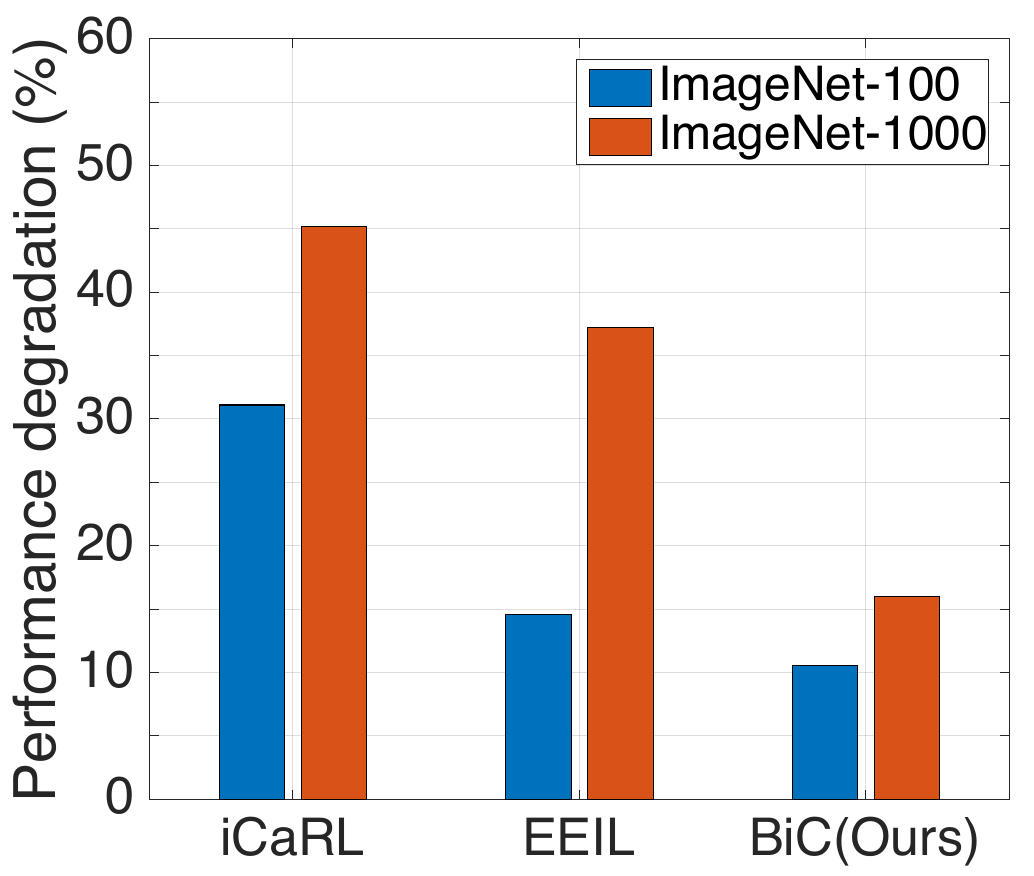}
		\end{center}
		\vspace{-4mm}
		\caption{Performance degradation of incremental learning algorithms on ImageNet-100 (100 classes) and ImageNet-1000 (1000 classes). Each dataset has 10 incremental steps. The degradation is the gap between the accuracy of the final incremental step and the accuracy of a non-incremental classifier, which is trained using all data. When the scale goes up (from ImageNet-100 to ImageNet-1000), the degradation for the state-of-the-art algorithms (iCaRL \cite{rebuffi2016icarl} and EEIL \cite{Castro_2018_ECCV}) increases. The degradation for our BiC method is small for both scales. Although iCaRL has similar relative degratation with our method (increase by 50\% from ImageNet-100 to ImageNet-1000), it performs poorly across the scales.}
		\label{fig:intro-curve-error}
		\vspace{-4mm}
	\end{figure}

	
	The missing data for old classes introduce two challenges - (a) maintaining the classification performance on old classes, and (b) balancing between old classes and new classes. 
	Distillation \cite{li2016learning, rebuffi2016icarl, Castro_2018_ECCV} has been used to effectively address the former challenge. Recent studies \cite{rebuffi2016icarl, Castro_2018_ECCV} also show that selecting a few exemplars from the old classes can alleviate the imbalance problem. These methods perform well on small datasets. However, they suffer from a significant performance degradation when the number of classes becomes large (e.g. thousands of classes). Fig. \ref{fig:intro-curve-error} demonstrates the performance degradation of these state-of-the-art algorithms, using a non-incremental classifier as the reference. When the number of classes increases from 100 to 1000, both iCaRL \cite{rebuffi2016icarl} and EEIL\cite{Castro_2018_ECCV} have more degradation. 
	
	
	\textit{Why is it more challenging to handle a large number of classes for incremental learning?} 
	We believe this is due to the coupling of two factors. First, the training data are unbalanced. Secondly, as the number of classes increases, it is more likely to have visually similar classes (e.g. multiple dog classes in ImageNet) across different incremental steps. Under the incremental constraint with data imbalance, the increasing number of visually similar classes is particularly challenging since the small margin around the boundary between classes is too sensitive to the data imbalance. The boundary is pushed to favor classes with more samples.
	
\begin{figure}[!t]
		\begin{center}
			\includegraphics[width=1.0\linewidth]{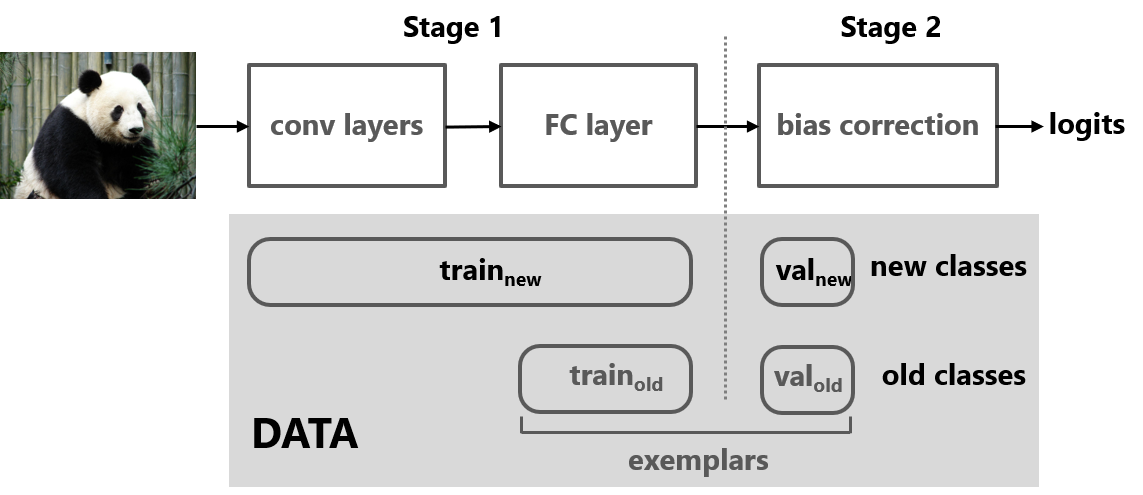}
		\end{center}
		\vspace{-4mm}
		\caption{Overview of our BiC method. The exemplars from the old classes and the samples of the new classes are split into training and validation sets. The training set is used to train the convolution layers and FC layer (in stage 1). The validation set is used for bias correction (in stage 2).}
		\label{fig:bic-overview}
		\vspace{-4mm}
	\end{figure}

	
	In this work, we present a method to address the data imbalance problem in large scale incremental learning. Firstly, we found a strong bias towards the new classes in the classifier layer (i.e. the last fully connected layer) of the convolution neural network (CNN).
	%
	%
	Based upon this finding, we propose a simple and effective method, called BiC (bias correction), to correct the bias. We add a bias correction layer after the last fully connected (FC) layer (shown in Fig. \ref{fig:bic-overview}), which is a simple linear model with two parameters. The bias correction layer is learned at the second stage, after learning the convolution layers and FC layer at the first stage. The data, including exemplars from the old classes and samples from the new classes, are split into a training set for the first stage and a validation set for the second stage. The validation set is helpful to approximate the real distribution of both old and new classes in the feature space, allowing us to estimate the bias in FC layer. We found that the bias can be effectively corrected with a small validation set.
	
	
	Our BiC method achieves remarkably good performance, especially on large scale datasets. The experimental results show that our method outperforms state-of-the-art algorithms (iCaRL\cite{rebuffi2016icarl} and EEIL \cite{Castro_2018_ECCV}) on two large datasets (ImageNet ILSVRC 2012 and MS-Celeb-1M) by a large margin. Our BiC method gains 11.1\% on ImageNet and 13.2\% on MS-Celeb-1M, respectively.
	\section{Related Work}

	Incremental learning has been a long standing problem in machine learning~\cite{cauwenberghs2001incremental,polikar2001learn++,mensink2013distance,kuzborskij2013n}.  Before the deep learning took off, people had been developing incremental learning techniques by leveraging linear classifiers, ensemble of weak classifiers, nearest neighbor classifiers, etc. Recently, thanks to the exciting progress in deep learning, there has been a lot of research on incremental learning with deep neural network models. The work can be roughly divided into three categories depending on whether they require real data or synthetic data or nothing from the old classes. 
	
	\textbf{Without using old data:} Methods in the first category do not require any old data. \cite{jung2016less} presented a method for domain transfer learning. They try to maintain the performance on old tasks by freezing the final layer and discouraging the change of shared weights in feature extraction layers. \cite{kirkpatrick2017overcoming} proposed a technique to remember old tasks by constraining the important weights when optimizing a new task. One limitation of this approach is that the old and new tasks may conflict on these important weights. \cite{li2016learning} presented a method that applies knowledge distillation \cite{hinton2015distilling} to maintain the performance on old tasks.  \cite{li2016learning} separated the old and new tasks in multi-task learning, which is different from learning classifier incrementally. \cite{Shmelkov-et-al-ICCV-2017} applied knowledge distillation for learning object detectors incrementally. 
	\cite{rannen2017encoder} utilized autoencoder to retain the knowledge from old tasks.
	\cite{sun2018lifelong,sun2018active} updated knowledge dictionary for new tasks and kept dictionary coefficients for old tasks.
	
	\textbf{Using synthetic data:} Both \cite{shin2017continual} and \cite{venkatesan2017strategy} employed GAN \cite{goodfellow2014generative} to replay synthetic data for old tasks. \cite{shin2017continual}  applied cross entropy loss on synthesis data with the old solver's response as the target. \cite{venkatesan2017strategy} utilized a root mean-squared error for learning the response of old tasks on synthetic data. \cite{shin2017continual,venkatesan2017strategy} highly depends on the capability of generative models and struggles with complex objects and scenes.  
	
	\textbf{Using exemplars from old data:} Methods in the third category require part of the old data. \cite{rebuffi2016icarl} proposed a method to select a small number of exemplars from each old class. \cite{Castro_2018_ECCV} keeps classifiers for all incremental steps and used them as distillation. It introduces balanced fine-tuning and temporary distillation to alleviate the imbalance between the old and new classes. \cite{lopez2017gradient} proposed a continuous learning framework where the training samples for different tasks are used one by one during training. It constrains the cross entropy loss on softmax outputs of old tasks when the new task comes. \cite{xiao2014error} proposed a training method that grows a network hierarchically as new training data are added. Similarly, \cite{rusu2016progressive} increases the number of layers in the network to handle new coming data. 
	
	Our BiC method belongs to the third category, we keep exemplars from the old classes in the similar manner to \cite{rebuffi2016icarl, Castro_2018_ECCV}. However, we handle the data imbalance differently. We first locate a strong bias in the classifier layer (the last fully connected layer), and then apply a linear model to correct the bias using a small validation set. The validation set is a small subset of exemplars which is excluded from training and used for bias correction alone. Compared with the state of the art (\cite{rebuffi2016icarl, Castro_2018_ECCV}), our BiC method is more effective on large datasets with 1000+ classes.
	
	
	\section{Baseline: Incremental Learning using Knowledge Distillation} \label{sec:logits_def}
	\begin{figure}[!t]
		\begin{center}
			\includegraphics[width=\linewidth]{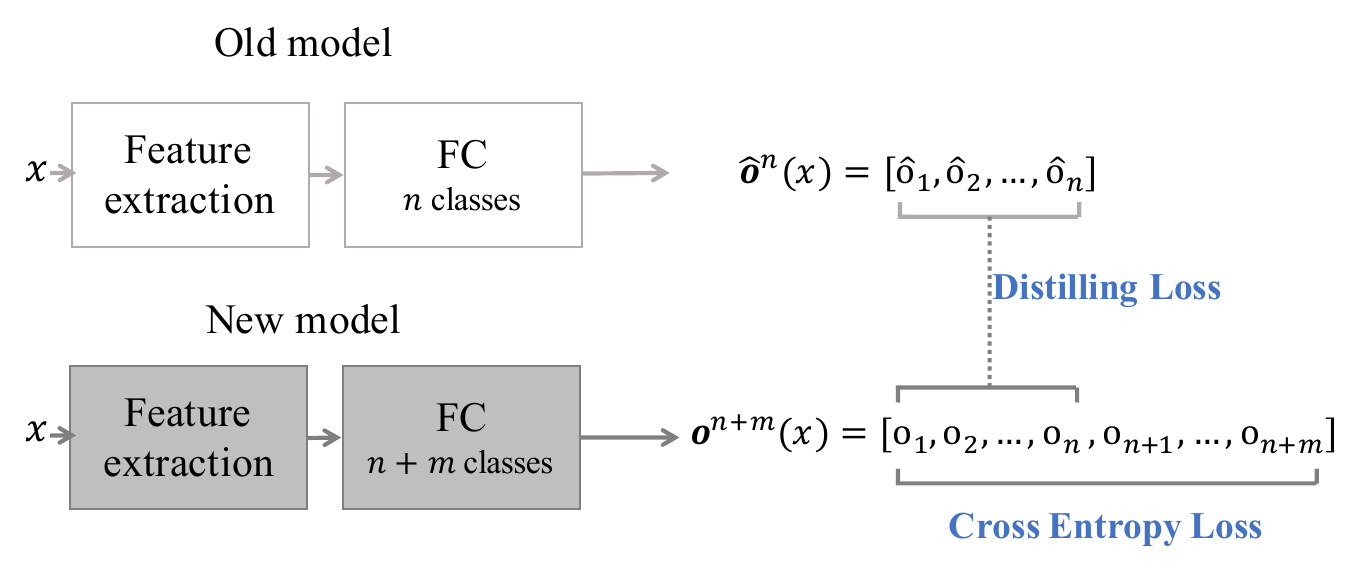}
		\end{center}
		\vspace{-4mm}
		\caption{Diagram of the baseline solution using distillation. It contains two losses: the distilling loss on old classes and the softmax cross-entropy loss on all old and new classes.}
		\label{fig:overview}
		\vspace{-4mm}
	\end{figure}
	
	In this section, we introduce a baseline solution for incremental learning using knowledge distillation \cite{li2016learning}. This is corresponding to the first stage in Fig. \ref{fig:bic-overview}. For an incremental step with $n$ old class and $m$ new classes, we learn a new model to perform classification on $n+m$ classes, by using the knowledge distillation from an old model that classifies the old $n$ classes (illustrated in Fig. \ref{fig:overview}). The new model is learned by using a distilling loss and a classification loss. 
	
	Let us denote the samples of the new classes as $X^m=\{(x_i, y_i), 1 \leq i \leq M, y_i \in [n+1,..,n+m]\}$, where $M$ is the number of new samples, $x_i$ and $y_i$ are the image and the label, respectively. The selected exemplars from the old $n$ classes are denoted as $\hat{X}^n=\{(\hat{x}_j, \hat{y}_j), 1 \leq j \leq N_s, \hat{y}_j \in [1,..,n]\}$, where $N_s$ is the number of selected old images ($N_s/n \ll M/m$).
	Let us also denote the output logits of the old and new classifiers as $\mathbf{\hat{o}}^n(x)=[\hat{o}_1(x),...,\hat{o}_n(x)]$ and $\mathbf{o}^{n+m}(x)=[o_1(x),...,o_n(x), o_{n+1}(x),...,o_{n+m}(x)]$ respectively. 
	The distilling loss is formulated as follows:
	 \begin{align}
	 	L_d &= \sum_{x\in \hat{X}^n  \cup X^m} \sum_{k=1}^n- \hat{\pi}_k(x) \log[\pi_k(x)], \label{eq:disloss} \\
	 	\hat{\pi}_k(x) &=\frac{e^{\hat{o}_k(x)/T}}{\sum_{j=1}^ne^{\hat{o}_j(x)/T}}, \quad \pi_k(x)=\frac{e^{o_k(x)/T}}{\sum_{j=1}^ne^{o_j(x)/T}}, \nonumber
	 \end{align}	
	 where 
	 $T$ is the temperature scalar. 
	  The distilling loss is computed for all samples from the new classes and exemplars from the old classes (i.e. $\hat{X}^n  \cup X^m$).
	
	We use the softmax cross entropy as the classification loss, which is computed as follows:
	\begin{align}
	L_c = \sum_{(x,y) \in \hat{X}^n \cup X^m} \sum_{k=1}^{n+m}-  \delta_{y=k} \log [p_k(x)],
	\label{eq:cls-loss}
	\end{align}
	where $ \delta_{y=k} $ is the indicator function and  $p_k(x)$ is the output probability (i.e. softmax of logits) of the $k$-th class in  $n+m$ old and new classes.
	
	The overall loss combines the distilling loss and the classification loss as follows:
	\begin{align}
	L = \lambda L_d + (1-\lambda)L_c, \label{eqloss}
	\end{align}
	where the scalar $\lambda$ is used to balance between the two terms. The scalar $\lambda$ is  set to $\frac{n}{n+m}$, where $n$ and $m$ are the number of old and new classes. 
	$\lambda$ is $0$ for the first batch since all classes are new. For the extreme case where $n\gg m$, $\lambda$ is nearly $1$, indicating the importance to maintain the old classes. 
	
	\section{Diagnosis: FC Layer is Biased}
	\begin{figure}[!t]
		\begin{center}
			\includegraphics[width=1.0\linewidth]{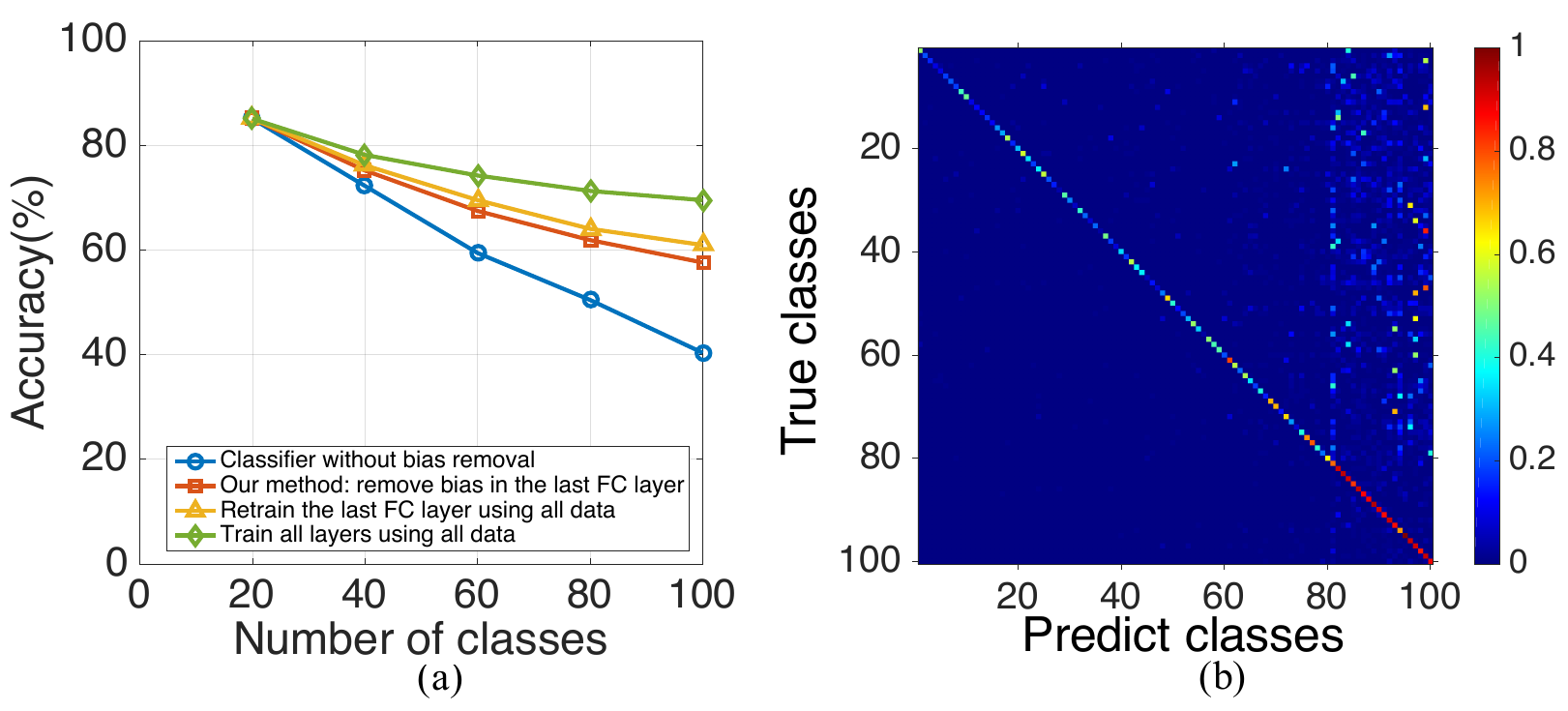}
		\end{center}
		\vspace{-4mm}
		\caption{ Experimental results on CIFAR-100 with split of 20 classes to validate the bias in the last FC layer.  (a) classification accuracy curves for baseline, our bias correction (BiC), retraining FC layer using all data, and training the whole network using all data (from bottom to top). (b) confusion matrix of the incremental classifier from 80 classes to 100 classes without bias removal. (Best viewed in color)}
		\label{fig:bias_FC}
		\vspace{-4mm}
	\end{figure}
	The baseline model has a bias towards the new classes, due to the imbalance between the number of samples from the new classes and the number of exemplars from the old classes. We have a hypothesis that \textit{the last fully connected layer is biased} as the weights are not shared across classes.
	To validate this hypothesis, we design an experiment on CIFAR-100 dataset with five incremental batches (each has 20 classes).
	
	First, we train a set of incremental classifiers using the baseline method. The classification accuracy quickly drops as more incremental steps arrive (shown as the bottom curve in Fig. \ref{fig:bias_FC}-(a)). For the last incremental step (class 81-100), we observe a strong bias towards the newest 20 classes in the confusion matrix (Fig.  \ref{fig:bias_FC}-(b)). 	
	Compared to the \textit{upper bound}, i.e. the classifiers learned using all training data (the top curve in Fig. \ref{fig:bias_FC}-(a)), the baseline model has a  performance degradation.
	
	
	Then, we conduct another experiment to evaluate if the fully connected layer is heavily biased. This experiment has two steps for each incremental batch: (a) applying the baseline model to learn both the feature and fully connected layers, (b) freezing the feature layers and retrain the fully connected layer alone using all training samples from both old and new classes. Compared to the baseline, the accuracy improves (the second top curve in Fig. \ref{fig:bias_FC}-(a)). The accuracy on the final classifier on 100 classes improves by 20\%. These results validate our hypothesis that the fully connected layer is heavily biased. 
	We also observe the gap between this result and the upper bound, which reflects the bias within the feature layers. In this paper, we focus on correcting the bias in the fully connected layer.
	

	\section{Bias Correction (BiC) Method \label{sec:bias-removal}}
	Based upon our finding that the fully connected layer is heavily biased, we propose a simple and effective bias correction method (BiC). Our method includes two stages in training (shown in Fig. \ref{fig:bic-overview}). Firstly, we train the convolution layers and the fully connected layer by following the baseline method. At the second stage, we freeze both the convolution and the fully connected layers, and estimate two bias parameters by using a small validation set. In this section, we discuss how the validation set is generated and the details of the bias correction layer.

	\subsection{Validation Set \label{sec:val-set}} 
	\begin{figure}[!t]
		\begin{center}
			\includegraphics[width=0.9\linewidth]{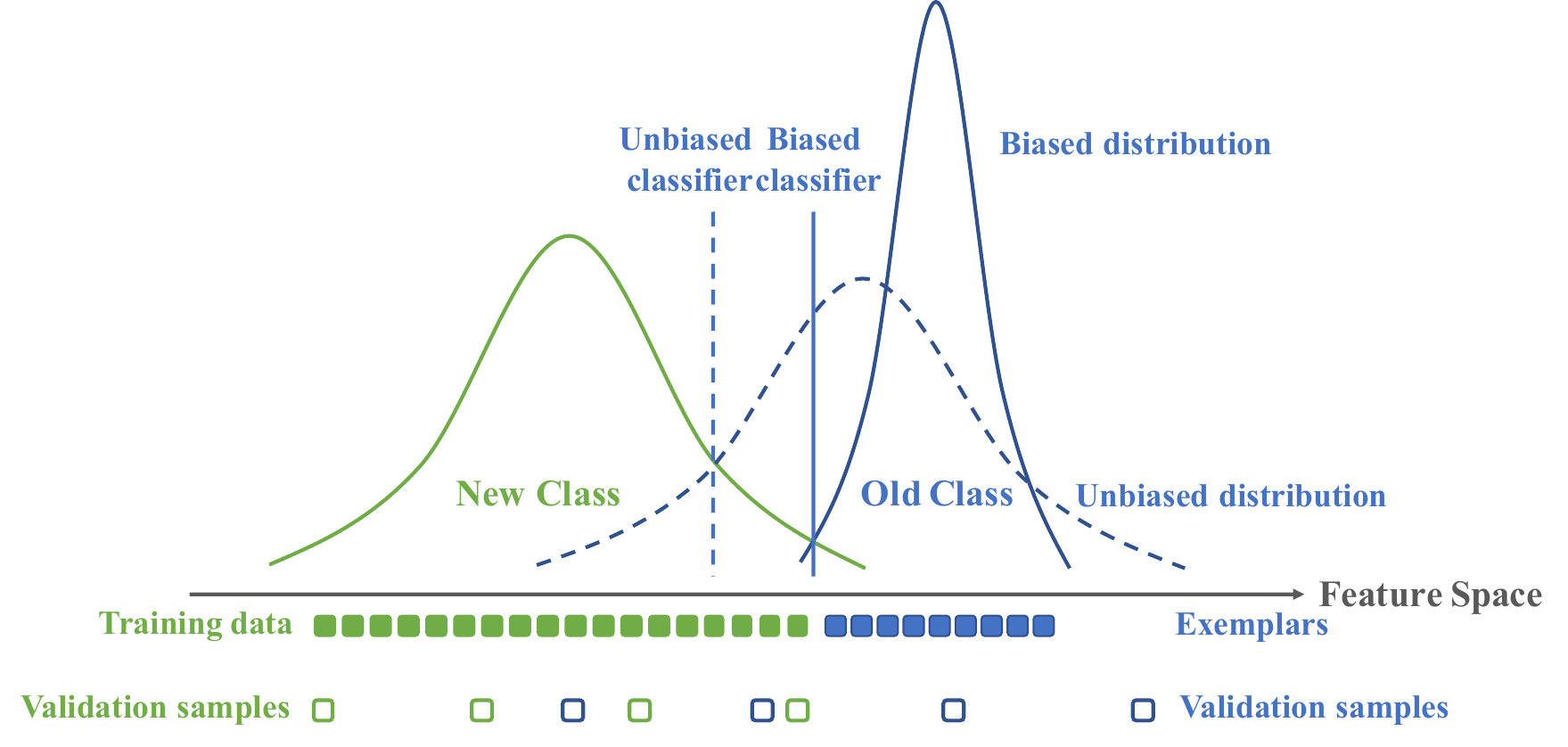}
		\end{center}
		\vspace{-4mm}
		\caption{Diagram of bias correction. Since the number of exemplars from old classes is small, they have narrow distributions on the feature space. This causes the learned classifier to prefer new classes.
			Validation samples, not involved in training feature representation, may better reflect the unbiased distribution of both old and new classes in the feature space. Thus, we can use the validation samples to correct the bias. (Best viewed in color)}
		\vspace{-4mm}
		\label{fig:biasdia}
	\end{figure}
	
	We estimate the bias by using a small validation set. The basic idea is to exclude the validation set from training the feature representation, allowing them to reflect the unbiased distribution of both old and new classes on the feature space (shown in Fig. \ref{fig:biasdia}). Therefore, we split the exemplars from the old classes and the samples from the new classes into a training set and a validation set. The training set is used to learn the convolution and fully connected layers (see Fig. \ref{fig:bic-overview}), while the validation set is used for the bias correction.
	
	
	Fig. \ref{fig:bic-overview} illustrates the generation of the validation set. The stored exemplars from the old classes are split into a training subset (referred to $train_{old}$) and a validation subset (referred to $val_{old}$). The samples for the new classes are also split into a training subset (referred to $train_{new}$) and a validation subset (referred to $val_{new}$). $train_{old}$ and $train_{new}$ are used to learn the convolution and FC layers (see Fig. \ref{fig:bic-overview}). $val_{old}$ and $val_{new}$ are used to estimate the parameters in the bias correction layer. Note that $val_{old}$ and $val_{new}$ are balanced. 
	
	\subsection{Bias Correction Layer} 
	The bias correction layer should be simple with a small number of parameters, since $val_{old}$ and $val_{new}$ have small size. Thus, we use a linear model (with two parameters) to correct the bias. This is achieved by adding a bias correction layer in the network (shown in Fig. \ref{fig:bic-overview}). 
	 We keep the output logits for the old classes ($1, \dots, n$) and apply a linear model to correct the bias on the output logits for the new classes ($n+1, \dots, n+m$) as follows: 
	\begin{align}
	q_k = \left \{ 
	\begin{aligned}
	& o_k & 1 \leq k \leq n &\\
	& \alpha o_k+\beta & n+1 \leq k \leq n+m &
	\end{aligned}
	\right. ,
	\label{pred}
	\end{align}
	where $\alpha$ and $\beta$ are the bias parameters on the new classes and $o_k$ (defined in Section \ref{sec:logits_def}) is the output logits for the $k$-th class. Note that the bias parameters ($\alpha$, $\beta$) are shared by all new classes, allowing us to estimate them with a small validation set. When optimizing the bias parameters, the convolution and fully connected layers are frozen. The classification loss (softmax with cross entropy) is used to optimize the bias parameters as follows:
	\begin{align}
	L_b = &-\sum_{k=1}^{n+m}  \delta_{y=k} \log [softmax(q_k)].
	\end{align}
	We found that this simple linear model is effective to correct the bias introduced in the fully connected layer.
	\section{Experiments}
	We compare our BiC method to the state-of-the-art methods on two large datasets (ImageNet ILSVRC 2012 \cite{russakovsky2015imagenet} and MS-Celeb-1M \cite{guo2016msceleb}), and one small dataset (CIFAR-100 \cite{krizhevsky2009learning}). We also perform ablation experiments to analyze different components of our approach. 
	\subsection{Datasets}
	We use all data in CIFAR-100 and ImageNet ILSVRC 2012 (referred to \textit{ImageNet-1000}), and randomly choose 10000 classes in MS-Celeb-1M (referred to \textit{Celeb-10000}). We follow iCaRL benchmark protocol \cite{rebuffi2016icarl} to select exemplars. The total number of exemplars for the old classes are fixed. The details of these three datasets are as follows:
	\\
	\textbf{CIFAR-100}: contains 60k $32 \times 32$ RGB images of 100 object classes. Each class has 500 training images and 100 testing images. 100 classes are split into 5, 10, 20 and 50 incremental batches.  2,000 samples are stored as exemplars. 
	\\
	\textbf{ImageNet-1000}: includes 1,281,167 images for training and 50,000 images for validation. 1000 classes are split into 10 incremental batches. 20,000 samples are stored as exemplars. 
	\\
	\textbf{Celeb-10000}: a random subset of 10,000 classes are selected from MS-Celeb-1M-base \cite{GuoZ17aa} face dataset which has 20,000 classes. MS-Celeb-1M-base is a smaller yet nearly noise-free version of MS-Celeb-1M \cite{guo2016msceleb}, which has near 100,000 classes with a total of 1.2 million aligned face images. For the randomly selected 10,000 classes, there are 293,052 images for training and 141,984 images for validation. 10000 classes are split into 10 incremental batches (1000 classes per batch). 50,000 samples are stored as exemplars. 
	
	For our BiC method, the ratio of train/validation split on the exemplars is 9:1 for CIFAR-100 and ImageNet-1000. This ratio is obtained from the ablation study (see Section \ref{secratioval}). We change the split ratio to 4:1 on Celeb-10000, allowing at least one validation image kept per person.
	
	\subsection{Implementation Details  }
	Our implementation uses TensorFlow \cite{abadi2016tensorflow}. We use an 18-layer ResNet \cite{he2016deep} for ImageNet-1000 and Celeb-10000 and use a 32-layer ResNet for CIFAR-100. The ResNet implementation is from TensorFlow official models\footnote{\url{https://github.com/tensorflow/models/tree/master/official/resnet}}.  The training details for each dataset are listed as follows:\\
	\textbf{ImageNet-1000} and \textbf{Celeb-10000}: Each incremental training has 100 epochs. The learning rate is set to 0.1 and reduces to 1/10 of the previous learning rate after 30, 60, 80 and 90 epochs.  The weight decay is set to 0.0001 and the batch size is 256. Image pre-processing follows the VGG pre-processing steps \cite{simonyan2014very}, including random cropping, horizontal flip and aspect preserving resizing and mean subtraction.  \\
	\textbf{CIFAR-100:} Each incremental training has 250 epochs. The learning rate starts from 0.1 initially and reduces to 0.01, 0.001 and 0.0001 after 100, 150 and 200 epochs, respectively. The weight decay is set to 0.0002 and the batch size is 128. Random cropping and horizontal flip is adapted for data augmentation following the original ResNet implementation \cite{he2016deep}. 
	
	For a fair comparison with iCaRL \cite{rebuffi2016icarl} and EEIL \cite{Castro_2018_ECCV}, we use the same networks, keep the same number of exemplars and follow the same protocols of splitting classes into incremental batches. We use the identical class order generated from iCaRL implementation\footnoteref{urlicarl} for CIFAR-100 and ImageNet-1000. On Celeb-10000, the class order is randomly generated and identical for all comparisons. The temperature scalar $T$ in Eq. \ref{eq:disloss} is set to 2 by following \cite{li2016learning,Castro_2018_ECCV}. 
	\subsection{Comparison on Large Datasets }\label{sec:largescale}
	
	\begin{table*}[!t]
		\begin{center}
			\begin{tabular}{|l|c|c|c|c|c|c|c|c|c|c|c|c|}
				\hline
				& 100 &  200 & 300 &   400 &  500 & 600 &  700 & 800 &   900 &  1000  \\
				\hline
				LwF \cite{li2016learning} & 90.0 & 77.0 & 68.0 &59.5& 52.5 & 49.5 & 46.5 & 43.0 & 40.5 & 39.0 \\
				iCaRL \cite{rebuffi2016icarl} & 90.0 &	83.0 & 77.5	&	70.5 &	63.0 & 57.5	& 53.5 & 50.0 & 48.0 & 44.0 \\
				EEIL \cite{Castro_2018_ECCV} & \textbf{95.0} &\textbf{95.5}	& 86.0	&	77.5&71.0	&68.0	&62.0	&59.8&55.0 & 52.0\\
				BiC(Ours) &{94.1} &	{92.5} &		\textbf{89.6}	&	\textbf{89.1}	&	\textbf{85.7} &	\textbf{83.2} &		\textbf{80.2} &		\textbf{77.5} &		\textbf{75.0}	& 	\textbf{73.2} \\
				\hline
			\end{tabular}
		\end{center}
		\caption{Incremental learning results (accuracy \%) on ImageNet-1000 dataset with an increment of 100 classes. LwF \cite{li2016learning} does not use any exemplars from the old classes. iCaRL \cite{rebuffi2016icarl}, EEIL \cite{Castro_2018_ECCV} and our BiC method use the same amount of exemplars from the old classes. Note that the numbers for LwF, iCaRL and EEIL on ImageNet-1000 are estimated from the figures in the original papers. The best results are marked in bold.
			\label{table:imagenet-1000}
		}
	\end{table*}
	\begin{table*}[!t]
		\begin{center}
			\begin{tabular}{|l|c|c|c|c|c|c|c|c|c|c|c|c|}
				\hline
				& 1000 &  2000 & 3000 &   4000 &  5000 & 6000 &  7000 & 8000 &   9000 &  10000  \\
				\hline
				iCaRL \cite{rebuffi2016icarl} & 94.31&	94.26&	91.09&	86.88&	81.06&	77.45&	75.29& 71.34& 68.78 & 65.56\\
				BiC(Ours) &\textbf{95.90} &	\textbf{96.65} &		\textbf{96.68}	&	\textbf{96.16}	&	\textbf{95.43} &	\textbf{	94.45} &		\textbf{93.35} &		\textbf{91.90} &		\textbf{90.18}	& 	\textbf{87.98} \\
				\hline
			\end{tabular}
		\end{center}
		\vspace{-2mm}
		\caption{Incremental learning results (accuracy \%) on Celeb-10000 dataset with an increment of 1000 classes. iCaRL \cite{rebuffi2016icarl} and our BiC method use the same amount of exemplars from the old classes. The best results are marked in bold.}
		\vspace{-2mm}
		\label{table:celeb-10000}
	\end{table*}
	
	In this section, we compare our BiC method with the state-of-the-art methods on two large datasets (ImageNet-1000 and Celeb-10000). The state-of-the-art methods include LwF \cite{li2016learning}, iCaRL\cite{rebuffi2016icarl} and EEIL \cite{Castro_2018_ECCV}. All of them utilize knowledge distillation to prevent catastrophic forgetting. iCaRL and EEIL keep exemplars for old classes, while LwF does not use any old data. 

\begin{figure}[!t]
		\begin{center}
			\includegraphics[width=0.9\linewidth]{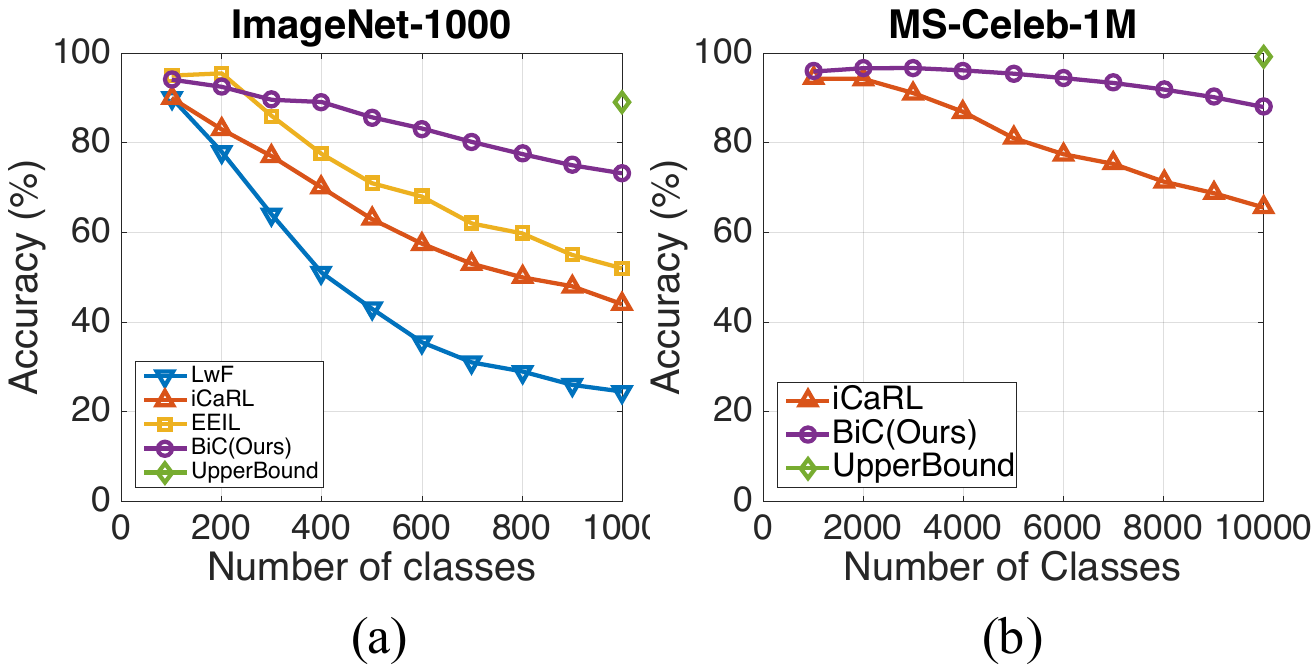}
		\end{center}
		\vspace{-4mm}
		\caption{Incremental learning results (accuracy \%) on (a) ImageNet-1000 and (b) Celeb-10000. Both datasets have ten incremental batches. The \textbf{Upper Bound} result, shown in the last step, is obtained by training a non-incremental model using all training samples from all classes. (Best viewed in color)}
		\vspace{-2mm}
		\label{fig:intro-curve}
	\end{figure}

	The incremental learning results on  ImageNet-1000 are shown in Table \ref{table:imagenet-1000} and Figure \ref{fig:intro-curve}-(a). Our BiC method outperforms both EEIL \cite{Castro_2018_ECCV} and  iCaRL \cite{rebuffi2016icarl} by a large margin.
	BiC has a small gain for the first couple of incremental batches compared with iCaRL and is worse than EEIL in the first two increments. However, the gain of BiC increases as more incremental batches arrive. Regarding the final incremental classifier on all classes, our BiC method outperforms EEIL \cite{Castro_2018_ECCV} and iCaRL \cite{rebuffi2016icarl} by 18.5\% and 26.5\% respectively. On average over 10 incremental batches, BiC outperforms EEIL \cite{Castro_2018_ECCV} and iCaRL \cite{rebuffi2016icarl} by 11.1\% and 19.7\% respectively.
	
	Note that the data imbalance increases as more incremental steps arrive. The reason is that the number of exemplars per old class decreases as the incremental step increases, since the total number of exemplars is fixed (by following the fix memory protocol in EEIL \cite{Castro_2018_ECCV} and iCaRL \cite{rebuffi2016icarl}). The gap between our BiC method and other methods becomes wider as the incremental step increases with more data imbalance. This demonstrates the advantage of our BiC method. 

	We also observe that EEIL performs better for the second batch (even higher than the first batch) on ImageNet-1000. 
	This is mostly due to the enhanced data augmentation (EDA) in EEIL that is more effective for the first couple of incremental batches when data imbalance is mild. 
	EDA includes random brightness shift, contrast normalization, random cropping and horizontal flipping. 
	In contrast, BiC only applies  random cropping and horizontal flipping.
	EEIL \cite{Castro_2018_ECCV} shows that EDA is effective for early incremental batches when data imbalance is not severe.
	Even without the enhanced data augmentation, our BiC still outperforms EEIL by a large margin on ImageNet-1000 starting from the third batch. 
	
	
	The incremental learning results on Celeb-10000 are shown in Table \ref{table:celeb-10000} and Figure \ref{fig:intro-curve}-(b). To the best of our knowledge, we have not seen any incremental learning method reporting results on 10,000 or more classes. The results for iCaRL is generated by applying its github implementation\footnote{\label{urlicarl}\url{https://github.com/srebuffi/iCaRL}}  on Celeb-10000 dataset. 
	For the first couple of incremental steps, our BiC method is slightly better than ($< 3\%$) iCaRL. But since the third incremental step, the gap becomes wider. At the last incremental step, BiC outperforms iCaRL by 22.4\%. The average gain over 10 incremental batches is 13.2\%.
	
	These results demonstrate our BiC method is more effective and robust to deal with a large number of classes. 
	As the number of classes increases, it is more frequent to have visually similar classes across different increment batches with unbalanced data. This introduces a strong bias towards new classes and misclassifies the old classes that are visually similar. Our BiC method is able to effectively reduce this bias and improve the classification accuracy.

	\subsection{Comparison between Different Scales}
	In this section, we compare our BiC method with the state-of-the-art on two different scales on ImageNet. The small scale deals with random selected 100 classes (referred to \textit{ImageNet-100}), while the large scale involves all 1000 classes (referred to \textit{ImageNet-1000}). Both scales have 10 incremental batches. This follows the same protocol with EEIL \cite{Castro_2018_ECCV} and  iCaRL \cite{rebuffi2016icarl}. The results for ImageNet-1000 is the same as in the previous section.
	
\begin{figure}[!t]
		\begin{center}
			\includegraphics[width=0.9\linewidth]{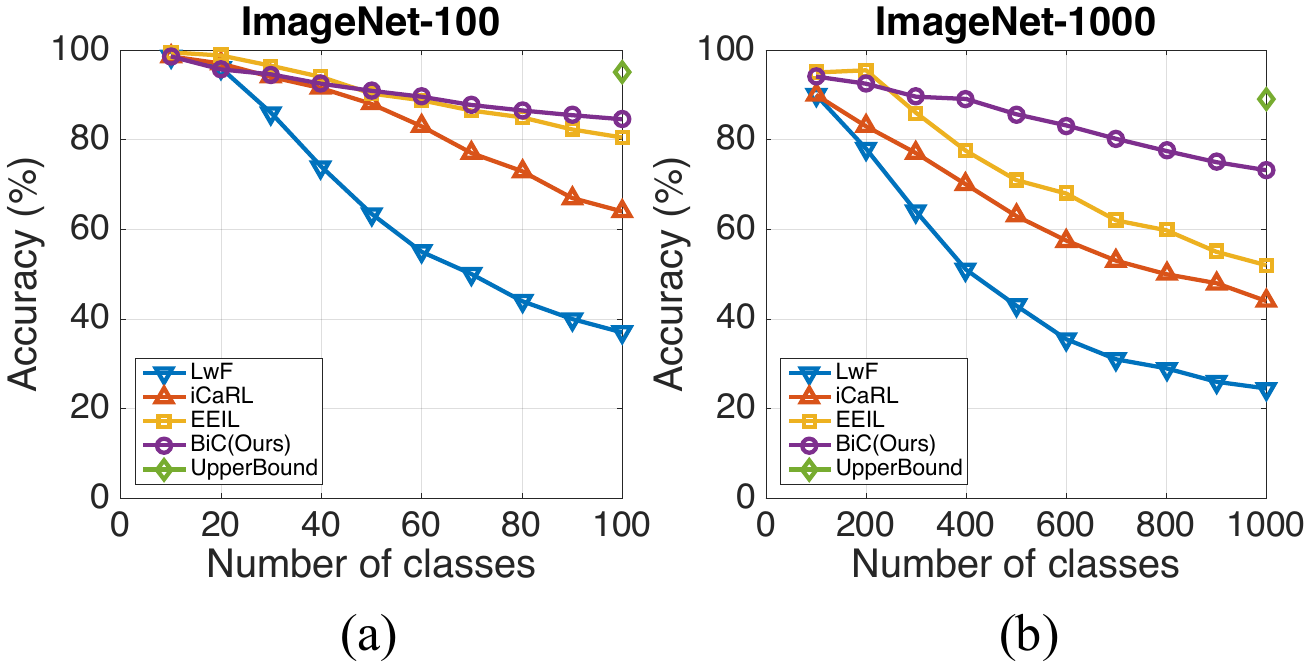}
		\end{center}\vspace{-4mm}
		\caption{Incremental learning results (accuracy \%) on ImageNet-100 and ImageNet-1000. Both have ten incremental batches. The \textbf{Upper Bound} result, shown in the last step, is obtained by training a non-incremental model using all training samples from all classes. (Best viewed in color)}
		\label{figimagenet}
		\vspace{-4mm}
	\end{figure}

	The incremental learning results on Imagenet-100 and ImageNet-1000 are shown in Fig. \ref{figimagenet}. 
	Our BiC method outperforms the state-of-the-art for both scales in terms of the final incremental accuracy and the average incremental accuracy. But the gain for the large scale is bigger. We also compare the final incremental accuracy (the last step) to the \textit{upper bound}, which is obtained by training a non-incremental model using all classes and their training data (shown at the last step in Fig. \ref{figimagenet}). Compared to the upper bound, our BiC method degrades 10.5\% and 16.0\% on ImageNet-100 and ImageNet-1000 respectively. However, EEIL \cite{Castro_2018_ECCV} degrades 15.1\% and 37.2\% and  iCaRL \cite{rebuffi2016icarl} degrades 31.1\% and 45.2\%. Compared with EEIL \cite{Castro_2018_ECCV} and  iCaRL \cite{rebuffi2016icarl}, which have more performance degradation from the small scale to large scale, our BiC method is much more consistent. This demonstrates that BiC has better capability to handle the large scale.
	
	We are aware that BiC is behind EEIL \cite{Castro_2018_ECCV} for the first three incremental batches on ImageNet-100. As explained in Section \ref{sec:largescale}, this is mostly due to enhanced data argumentation (EDA) in EEIL \cite{Castro_2018_ECCV}. 
	
	
	\subsection{Comparison on a Small Dataset}
	\begin{figure*}[!t]
		\begin{center}
			\includegraphics[width=0.95\linewidth]{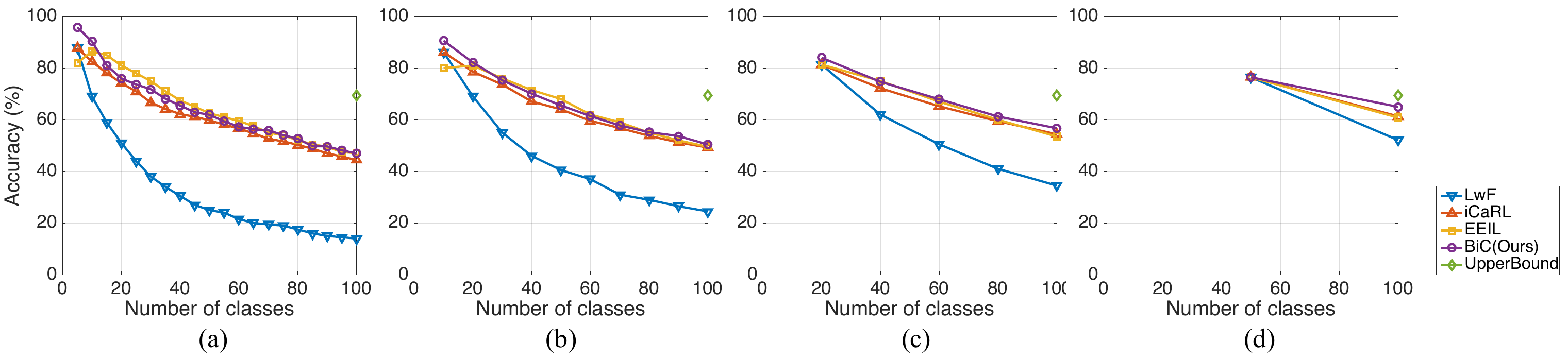}		
		\end{center}\vspace{-4mm}
		\caption{Incremental learning results on CIFAR-100 with split of (a) 5 classes, (b) 10 classes, (c) 20 classes and (d) 50 classes. The \textbf{Upper Bound} result, shown in the last step, is obtained by training a non-incremental model using all training samples for all classes. (Best viewed in color)}
		\label{figcifar}\vspace{-2mm}
	\end{figure*}
	
	We also compare our BiC method with the state-of-the-art algorithms on a small dataset - CIFAR-100 \cite{krizhevsky2009learning}. The incremental learning results with four different splits of 5, 10, 20 and 50 classes are shown in Fig. \ref{figcifar}. Our BiC method has similar performance with iCaRL \cite{rebuffi2016icarl} and EEIL \cite{Castro_2018_ECCV}. BiC is better on the split of 50 and 20 classes, but is slightly behind EEIL on the split of 10 and 5 classes. The margins are small for all splits. 

	Although our method focuses on the large scale incremental learning, it is also compelling on the small scale. Note that EEIL has more data augmentation such as brightness augmentation and contrast normalization, which are not utilized in LwF, iCaRL or BiC. 
	
	
	\subsection{Ablation Study}
	We now analyze the components of our BiC method and demonstrate their impact. The ablation study is performed on CIFAR-100 \cite{krizhevsky2009learning}, as incremental learning on large dataset is time consuming.  The ablation study is performed on CIFAR-100 with an incremental of 20 classes. The size of the stored exemplars from old classes is 2,000. 
	%
	In the following ablation study, we analyze (a) the impact of bias correction, (b) the split of validation set, and (c) the sensitivity of exemplar selection.

	\textbf{The Impact of Bias Correction}
	\begin{table*}[!t]
	
		\begin{center}
			\scalebox{1}{
				\begin{tabular}{|l|c|c|c|c|c|c|c|c|c|}
					\hline
					Variations & cls loss & distilling loss & bias removal & FC retrain & 20 &  40 & 60 &   80 &  100\\
					\hline\hline
					baseline-1& \checkmark & & & & 84.40&	68.30&55.10&48.52&39.83\\
					baseline-2 & \checkmark & \checkmark & & & \textbf{85.05}&	 72.22 &59.41& 50.43 & 40.34\\
					BiC(Ours)  & \checkmark & \checkmark &  \checkmark & & 84.00&	\textbf{74.69} & \textbf{67.93} & \textbf{61.25}& \textbf{56.69}\\
					\hline
					upper bound & \checkmark& \checkmark& &\checkmark & 84.39& 76.15& 69.51 & 64.03 &  60.93\\ 
					\hline
				\end{tabular}
			}
		\end{center}
		 \vspace{-2mm}	
		 \caption{Incremental learning results on CIFAR-100 with a batch of 20 classes. baseline-1 uses the classification loss alone. baseline-2 uses both the distilling loss and the classification loss. BiC corrects the bias in FC layer of baseline-2. Upper bound retrains the last FC layer using all samples from both old and new classes after learning the model of baseline-2.
		 The best results are marked in bold.\label{tableicarl20ana}} 
		\vspace{-2mm}	
	\end{table*}
	We compare our BiC method with two variations of baselines and the upper bound, to analyze the impact of bias correction. The baselines and the upper bound are explained as follows:\\
	\textbf{baseline-1:} the model is trained using the classification loss alone (Eq. \ref{eq:cls-loss}). \\
	\textbf{baseline-2:} the model is trained using both the distilling loss and the classification loss (Eq. \ref{eqloss}). Compared to the baseline-1, the distilling loss is added. \\
	\textbf{BiC:} the model is trained using both the distilling loss and the classification loss, with the bias correction.\\
	\textbf{upper bound:} the model is firstly trained using both the distilling loss and classification loss. Then, the feature layers are frozen and the classifier layer (i.e. the fully connected layer) is retrained using all training data (including the samples from the old classes that are not stored). Although it is infeasible to have all training samples from the old classes, it shows the upper bound for the bias correction in the fully connected layer.\\
	
	The incremental learning results are shown in Table \ref{tableicarl20ana}. With the help of the knowledge distillation, baseline-2 is slightly better than baseline-1 since it retains the classification capability on the old classes. However, both baseline-1 and baseline-2 have low accuracy on the final step to classify all 100 classes (about 40\%). This is mainly because of the data imbalance between the old and new classes. When using the bias correction, BiC improves the accuracy on all incremental steps. The classification accuracy on the final step (100 classes) is boosted from 40.34\% to 56.69\%. This demonstrates that the bias is a big issue and our method is effective to address it. Furthermore, our method is close to the upper bound. The small gap (4.24\%) from our approach 56.69\% to the upper bound 60.93\% shows the superiority of our method.

	\begin{figure*}[!t]
		\begin{center}
			\includegraphics[width=0.95\linewidth]{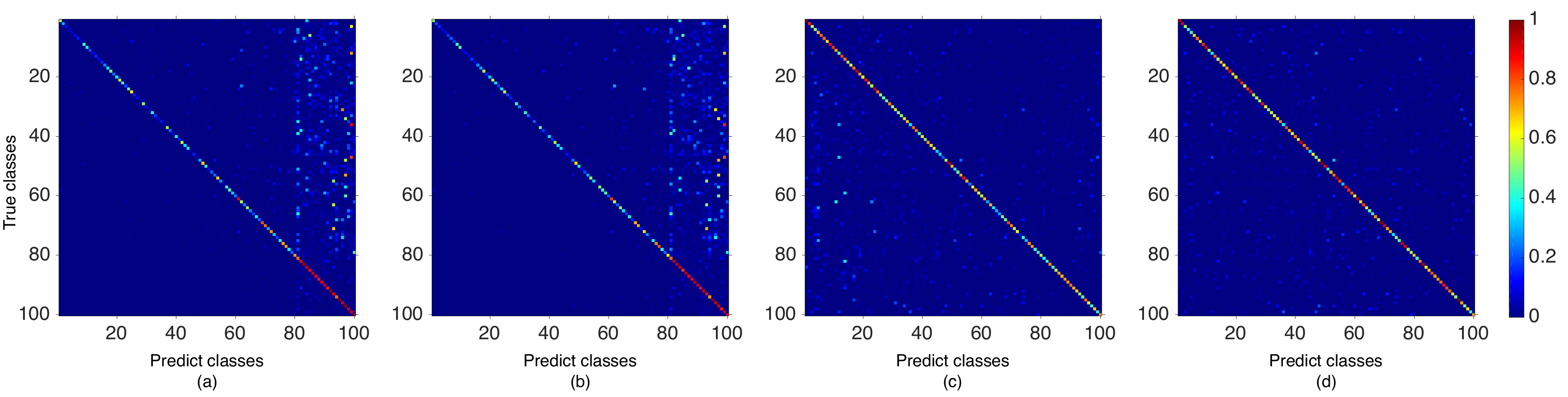}
		\end{center}\vspace{-2mm}
		\caption{Confusion matrices of four different variations: (a) baseline-1 (b) baseline-2,  (c) BiC,  (d) upper bound. Both baseline-1 and baseline-2 have strong bias towards new classes. BiC is capable to remove most of the bias and have similar confusion matrix with the upper bound. (Best viewed in color)}
		\label{confusionallana}\vspace{-4mm}
	\end{figure*}

	The confusion matrices of these four variations are shown in Fig. \ref{confusionallana}. Clearly, baseline-1 and baseline-2 suffer from the bias towards the new classes (strong confusions on the last 20 classes). BiC reduces the bias and has similar confusion matrix to the upper bound.
	
	These results validate our hypothesis that there exists a strong bias towards the new classes in the last fully connected layer. In addition, the results demonstrate that the proposed bias correction using a linear model on a small validation set is capable to correct the bias.
	
	\begin{table}[!t]
		\begin{center}
			\begin{tabular}{|c|c|c|c|c|c|}
				\hline
				$train_{old}$:$val_{old}$ & 20 &  40 & 60 &   80 &  100\\
				\hline
				9:1 &  84.00&	\textbf{74.69} & \textbf{67.93} & \textbf{61.25}& \textbf{56.69}\\
				\hline
				8:2 & 84.50  &	73.19& 65.01  & 58.68 & 54.31 \\
				\hline
				7:3 & \textbf{84.70}  &	71.60 & 63.68 & 58.12 & 53.74 \\
				\hline
				6:4 & 83.33  & 68.84  & 62.21  & 56.00 & 51.17  \\
				\hline
			\end{tabular}
		\end{center}
		\caption{Incremental learning results on CIFAR-100 with a batch of 20 classes for different training/validation split on exemplars from old classes. The training set is used to learn the feature and classifier layers, and the validation set is used to learn the bias correction layer. The best results are marked in bold. 
			\label{tableratio}}
		\vspace{-2mm}
	\end{table}
	
	\textbf{The Split of Validation Set}\label{secratioval}
	We study the impact of different splits of the validation set (see Section \ref{sec:val-set}). As illustrated in Fig. \ref{fig:bic-overview}, our BiC splits the stored exemplars from the old classes into a training set ($train_{old}$) and a validation set ($val_{old}$). The samples from the new classes also have a train/val split ($train_{new}$ and $val_{new}$). $train_{old}$ and $train_{new}$ are used to learn the convolution layers and the fully connected layer, while $val_{old}$ and $val_{new}$ are used to learn the bias correction layer. Note that $val_{old}$ and $val_{new}$ are balanced, having the same number of samples per class. Since only a few exemplars  (i.e. $train_{old}\bigcup val_{old}$) are stored for the old classes, it is critical to find a good split that deals with the trade-off between training the feature representation and correcting the bias in the fully connected layer.
	
	Table \ref{tableratio} shows the incremental learning results for four different splits of $train_{old}:val_{old}$. The split of 9:1 has the best classification accuracy for all four incremental steps. The column 20 refers to learning a classifier for the first 20 classes, without incremental learning. As the portion for the validation set increases, the performance drops consistently due to the lack of exemplars (from the old classes) to train the feature layers. A small validation set ($\frac{1}{10}$ of exemplars) is good enough to estimate the bias parameters ($\alpha$ and $\beta$ in Eq. \ref{pred}). In this paper, we use split 9:1 for all other experiments except Celeb-10000. The split 4:1 is adopted in Celeb-10000, as each old class only has 5 exemplars for the last incremental step.

	\textbf{The Sensitivity of Exemplar Selection}
	We also study the impact of different exemplar management strategies. We compare two strategies: (a) random selection, and (b) the exemplar management strategy proposed by iCaRL \cite{rebuffi2016icarl}. iCaRL maintains the samples that closed to the class center in the feature space. Both strategies store 2,000 exemplars from old classes. 
	The incremental learning results are shown in Table \ref{tableexemplar}. iCaRL exemplar management strategy performs slightly better than the random selection. The gap is about 1\%. This demonstrates that our method is not sensitive to the exemplar selection.  
	\begin{table}[!t]
		\begin{center}
			\begin{tabular}{|l|c|c|c|c|c|}
				\hline
				& 20 &  40 & 60 &   80 &  100\\
				\hline
				random & \textbf{85.20}  &	74.59 & 66.76 & 60.14 & 55.55 \\
				\hline
				iCaRL \cite{rebuffi2016icarl}&  84.00& \textbf{74.69} & \textbf{67.93} & \textbf{61.25} & \textbf{56.69} \\
				\hline
			\end{tabular}
		\end{center}
		\caption{Incremental learning results on CIFAR-100 with a batch of 20 classes for different exemplar management strategies. The best results are marked in bold. 
			\label{tableexemplar}}
		\vspace{-2mm}
	\end{table}
	\section{Conclusions}
	In this paper, we proposed a new method to address the imbalance issue in incremental learning, which is critical when the number of classes becomes large. Firstly, we validated  our hypothesis that the classifier layer (the last fully connected layer) has a strong bias towards the new classes, which has substantially more training data than the old classes. Secondly, we found that this bias can be effectively corrected by applying a linear model with a small validation set. Our method has excellent results on two large datasets with 1,000+ classes (ImageNet ILSVRC 2012 and MS-Celeb-1M), outperforming the state-of-the-art by a large margin (11.1\% on ImageNet ILSVRC 2012 and 13.2\% on MS-Celeb-1M). 
	\section{Acknowledgments}
	Part of the work was done when Yue Wu was an intern
	at Microsoft. This research is supported in part by the NSF IIS Award 1651902.
	
	{\small
		\bibliographystyle{ieee_fullname}
		\bibliography{egbib}
	}
	
\end{document}